\documentclass[letterpaper, 10 pt, conference]{ieeeconf}  

\IEEEoverridecommandlockouts                              

\usepackage{graphicx}
\usepackage{amsmath}
\usepackage{amssymb}
\usepackage{cite}
\usepackage{xcolor}
\usepackage{multirow}
\usepackage{booktabs}
\usepackage{array}
\usepackage{algorithm}
\usepackage[noend]{algpseudocode}
\usepackage{caption}
\usepackage{makecell}
\usepackage{pifont}
\usepackage{hyperref}
\usepackage{subcaption}
\newcommand{\mymat}[1]{\mathbf{#1}}

\newcommand{\timederiv}{\partial}
\newcommand{\optctrl}{CND}

\newcommand{\stxt}[1]{}
\newcommand{\vtxt}[1]{}

\newcommand*\colourcheck[1]{%
  \expandafter\newcommand\csname #1check\endcsname{\textcolor{#1}{\ding{52}}}%
}
\newcommand*\colourcross[1]{%
  \expandafter\newcommand\csname #1cross\endcsname{\textcolor{#1}{\ding{56}}}%
}
\colourcheck{green}
\colourcross{red}

\title{\LARGE \bf
Neural Optimal Control using Learned System Dynamics
}

\author{Selim Engin$^{1}$ and Volkan Isler$^{1}$
\thanks{This work was supported in part by NSF grants \#1617718 and \#2022894.}
\thanks{$^{1}$Authors are with Department of Computer Science and Engineering,
        University of Minnesota, 100 Union St SE, Minneapolis, MN 55455, USA
        {\tt\small \{engin003, isler\}@umn.edu}}
}

\begin{document}

\maketitle
\thispagestyle{empty}
\pagestyle{empty}

\begin{abstract}
We study the problem of generating control laws for systems with unknown dynamics.
Our approach is to represent the controller and the value function with neural networks, and to train them using loss functions adapted from the Hamilton-Jacobi-Bellman (HJB) equations.
In the absence of a known dynamics model, our method first learns the state transitions from data collected by interacting with the system in an offline process.
The learned transition function is then integrated to the HJB equations and used to forward simulate the control signals produced by our controller in a feedback loop.

In contrast to trajectory optimization methods that optimize the controller for a single initial state, our controller can generate near-optimal control signals for initial states from a large portion of the state space. Compared to recent model-based reinforcement learning algorithms, we show that our method is more sample efficient and trains faster by an order of magnitude.
We demonstrate our method in a number of tasks, including the control of a quadrotor with 12 state variables.
\end{abstract}


\newcommand{\approaches}{
\begin{table}[h]
\centering 
\begin{tabular}{l c c c} 
\toprule
Characteristics: & HJB-LSM  & NMPC & CT-MBRL \\ [0.5ex]
\hline 
Solution over the entire state space  & \greencheck & \redcross & \greencheck \\ 
No space discretization  & \redcross & \greencheck & \greencheck \\ 
No requirement of true dynamics  & \redcross & \redcross & \greencheck \\ 
Sample efficiency  & - & \greencheck & \redcross \\ 
\bottomrule
\end{tabular}
\caption{A comparison of various approaches to solve nonlinear optimal control problems, highlighting the differences in their characteristics. We compare a) numerical level-set methods for solving HJB equations (HJB-LSM), b) nonlinear model predictive control (NMPC) and c) continuous-time model-based reinforcement learning (CT-MBRL).}
\label{tab:approaches} 
\end{table}
}

\newcommand{\fklearning}{
\begin{table*}[!htbp]
  \centering
  \resizebox{\linewidth}{!}{%
  \begin{tabular}{l c c | c c  | c c | c c}
    \toprule
     & \multicolumn{2}{c}{Dubins Car (Dim: 3)} &  \multicolumn{2}{c}{Acrobot (Dim: 4)} & \multicolumn{2}{c}{Cartpole (Dim: 4)} & \multicolumn{2}{c}{Quadrotor (Dim: 12)} \\
    {\textbf{Strategies}} & Mean (s.d.) $\downarrow$ &  Median (iqr) $\downarrow$  & Mean (s.d.) $\downarrow$ &  Median (iqr) $\downarrow$ & Mean (s.d.) $\downarrow$ &  Median (iqr) $\downarrow$  & Mean (s.d.) $\downarrow$ &  Median (iqr) $\downarrow$ \\
    
    \hline
ReLU (w/o $\nabla f$) & 0.0071\scriptsize{$\pm$0.0033} & 0.0067\scriptsize{$\pm$0.0043} & 0.7717\scriptsize{$\pm$0.5395} & 0.6768\scriptsize{$\pm$0.5464} & 0.0819\scriptsize{$\pm$0.0551} & 0.0693\scriptsize{$\pm$0.0611} & 20.8816\scriptsize{$\pm$11.2258} & 18.6534\scriptsize{$\pm$13.0919} \\
ReLU (with $\nabla f$) & 0.0134\scriptsize{$\pm$0.0090} & 0.0113\scriptsize{$\pm$0.0094} & 0.6456\scriptsize{$\pm$0.4999} & 0.5432\scriptsize{$\pm$0.4726} & 0.0734\scriptsize{$\pm$0.0511} & 0.0621\scriptsize{$\pm$0.0543} & 35.9359\scriptsize{$\pm$18.6272} & 30.7491\scriptsize{$\pm$24.2143} \\
Tanh (w/o $\nabla f$) & 0.0017\scriptsize{$\pm$0.0009} & 0.0016\scriptsize{$\pm$0.0009} & 0.4328\scriptsize{$\pm$0.3702} & 0.3551\scriptsize{$\pm$0.2858} & 0.0210\scriptsize{$\pm$0.0111} & 0.0189\scriptsize{$\pm$0.0119} & 8.3696\scriptsize{$\pm$5.7249} & 6.8628\scriptsize{$\pm$5.9049} \\
Tanh (with $\nabla f$) & 0.0015\scriptsize{$\pm$0.0003} & 0.0015\scriptsize{$\pm$0.0005} & 0.2834\scriptsize{$\pm$0.1800} & 0.2488\scriptsize{$\pm$0.1958} & 0.0137\scriptsize{$\pm$0.0067} & 0.0124\scriptsize{$\pm$0.0079} & 18.6461\scriptsize{$\pm$8.2238} & 18.6983\scriptsize{$\pm$11.9576} \\
Sine (w/o $\nabla f$) & 0.0014\scriptsize{$\pm$0.0008} & 0.0013\scriptsize{$\pm$0.0008} & 0.0673\scriptsize{$\pm$0.0417} & \textbf{0.0585\scriptsize{$\pm$0.0414}} & 0.0149\scriptsize{$\pm$0.0066} & 0.0140\scriptsize{$\pm$0.0083} & \textbf{1.3109\scriptsize{$\pm$0.9937}} & \textbf{1.0362\scriptsize{$\pm$0.9085}} \\
Sine (with $\nabla f$) & \textbf{0.0008\scriptsize{$\pm$0.0003}} & \textbf{0.0008\scriptsize{$\pm$0.0004}} & \textbf{0.0664\scriptsize{$\pm$0.0358}} & 0.0595\scriptsize{$\pm$0.0419} & \textbf{0.0101\scriptsize{$\pm$0.0043}} & \textbf{0.0096\scriptsize{$\pm$0.0055}} & 15.6214\scriptsize{$\pm$8.1264} & 15.8460\scriptsize{$\pm$13.2569} \\
    \bottomrule
  \end{tabular}%
  }
  \caption{Comparison of different activations (ReLU, Tanh and Sine) and loss functions (supervised with $f$ and supervised with $f + \nabla f$) for learning the state transitions of the Dubins Car, Acrobot, Cartpole and Quadrotor systems. We report the mean ($\pm$ standard deviation) and median ($\pm$ interquantile range) errors of the predicted state transitions.}
  \label{table:fk_learning} \vspace{-0.15in}
\end{table*}}

    


\newcommand{\gymtasks}{
\begin{table}[h]
  \centering
  \resizebox{\linewidth}{!}{%
  \begin{tabular}{l c c c | c c c}
    \toprule
     & \multicolumn{3}{c}{Acrobot} & \multicolumn{3}{c}{Cartpole} \\
    {\textbf{Methods}} & Term. error &  Control magn. & NFE  & Term. error &  Control magn. & NFE \\
    
    \hline
    Deep PILCO~\cite{gal2016improving} & 0.284\scriptsize{$\pm$0.10} & 4.624\scriptsize{$\pm$2.63} & 4789 & 0.146\scriptsize{$\pm$0.20} & 1.168\scriptsize{$\pm$1.17} & 7057 \\
PETS~\cite{chua2018deep} & 3.095\scriptsize{$\pm$0.69} & 6.968\scriptsize{$\pm$0.36} & 12602 & 0.910\scriptsize{$\pm$0.22} & 1.311\scriptsize{$\pm$0.82} & 12602 \\
ENODE~\cite{yildiz2021continuous} & 1.133\scriptsize{$\pm$0.52} & 4.934\scriptsize{$\pm$1.83} & 27812  & 0.170\scriptsize{$\pm$0.10} & 1.317\scriptsize{$\pm$1.18} & 6262 \\
Ours & 1.071\scriptsize{$\pm$0.63} & 5.936\scriptsize{$\pm$0.72} & 1000 & 0.049\scriptsize{$\pm$0.03} & 0.742\scriptsize{$\pm$0.47} & 1000 \\
    \bottomrule
  \end{tabular}%
  }
  \caption{Comparison of methods to solve the optimal control problems Acrobot and Cartpole without knowing the dynamics beforehand. We report the mean terminal state error, mean control magnitude along the trajectories, and the number of function evaluations used to train the controllers.}
  \label{table:optctrl_gym} 
\end{table}}

\newcommand{\quadrotorctrl}{
\begin{table}[h]
  \centering
  \begin{tabular}{l c | c c}
    \toprule
    {\textbf{Methods}} & S.T. & Terminal error &  Control magn. \\
    \hline
    NeuralOC~\cite{onken2021neural} & $f$ & 0.4601\scriptsize{$\pm$0.1552} & 45.9289\scriptsize{$\pm$17.8367} \\
    Ours & $f$ & 0.0690\scriptsize{$\pm$0.2119}  & 60.6403\scriptsize{$\pm$39.7774}  \\
    Ours & $f_\theta$ & 0.5666\scriptsize{$\pm$0.1131}  & 71.3644\scriptsize{$\pm$29.4289}   \\
    \hline
    Ours ($R = 0$) & $f_\theta$ & 0.4161\scriptsize{$\pm$0.1732} & 72.4603\scriptsize{$\pm$18.6037}   \\
    \bottomrule
  \end{tabular}%
  \caption{Comparison of HJB-based neural controllers for controlling the quadrotor to reach a desired state. In terms of the terminal error, our method outperforms the baseline significantly when the state transitions $f$ is known. Our method performs competitively even when the system dynamics $f_\theta$ is learned and imperfect.}
  \label{table:quadrotor} 
\end{table}}

\newcommand{\dubinsctrl}{
\begin{table}[h]
  \centering
  \begin{tabular}{l | c c c}
    \toprule
    {\textbf{Methods}} & Trajectory len. &  Compute time (s) & Success rate \\
    \hline
    RRT$^*$  (iter: 100) & 4.8871\scriptsize{$\pm$0.9607}  & 6.9313  & 0.889 \\
    RRT$^*$  (iter: 200) &  4.7343\scriptsize{$\pm$0.8451} & 26.0595  & 0.939 \\
    Ours & 4.1076\scriptsize{$\pm$0.6916}  & 0.0634  & 0.999 \\
    \bottomrule
  \end{tabular}%
  \caption{Comparison of our method against RRT$^*$~\cite{karaman2011sampling}, a sampling-based motion planning algorithm, for controlling a Dubins vehicle to a goal state in the presence of an obstacle.}
  \label{table:optctrldubins} 
\end{table}}

\section{Introduction}
Many robotics tasks, such as navigation and locomotion, can be formulated as optimal control problems, which are challenging to solve especially when the system has complex dynamics and the state space is high dimensional. Optimal control formulations are interesting since they allow us to express tasks as simple cost functions and leave the rest to the control algorithm.

In this paper, we study nonlinear optimal control problems in finite time horizon.
One way to solve these problems is through trajectory optimization methods that use nonlinear programming, such as nonlinear model predictive control~\cite{johansen2011introduction}, which optimize the control signals to generate optimal trajectories given an initial state.
However, these methods are local, meaning that they perform optimization for a single initial state, and their recomputation for other initial states taking seconds might be too restrictive for real-time applications. Updating the initial state can be needed in a variety of robotics tasks under uncertainties, due to environment disturbances and sensor limitations, for example.

\begin{figure}[t]
    \centering
    \includegraphics[width=\columnwidth]{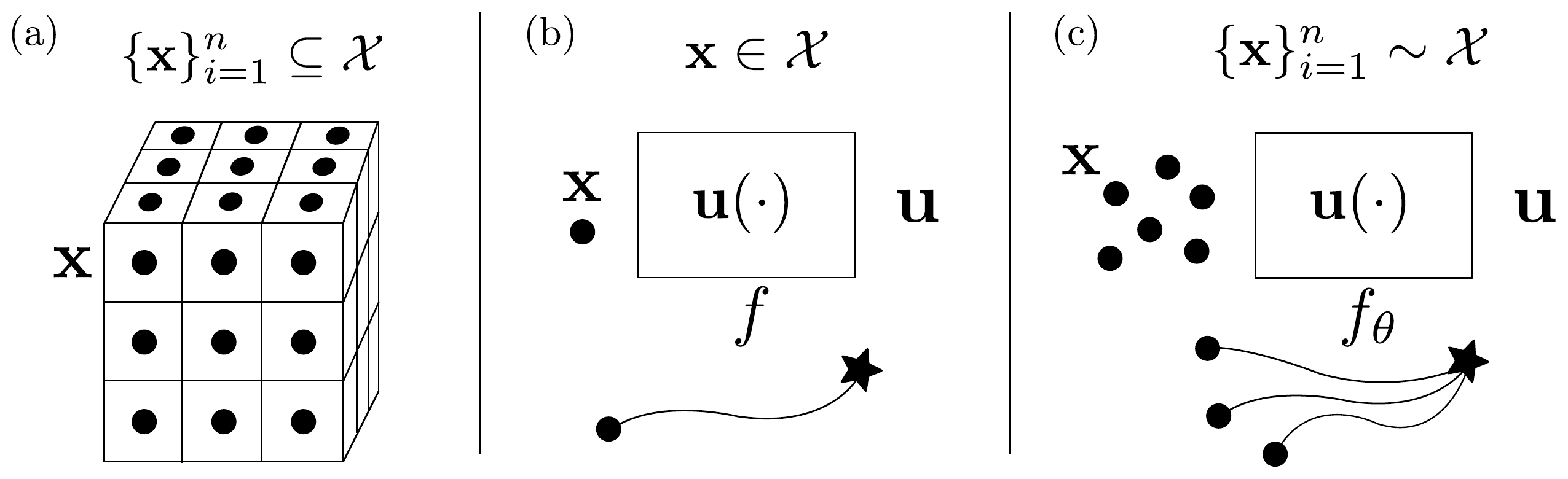}
    \caption{An overview of different approaches for solving nonlinear optimal control problems. \textbf{(a)} Approaches that solve the HJB equation with numerical methods offer global solutions, however are restricted to lower dimensions due to discretization of the state space $\mathcal{X}$. \textbf{(b)} Trajectory optimization methods optimize the controller $\mymat{u}(\cdot)$ for a single initial state to reach the target using known state transitions. 
    \textbf{(c)} Our approach generates a controller trained using learned transitions $f_\theta$, for initial states from a large portion of $\mathcal{X}$.}
    \label{fig:teaser} 
\end{figure}

Generating control signals in a global fashion is possible using the Hamilton-Jacobi-Bellman (HJB) equations, whose solution is the \emph{value function} or the optimal cost-to-go. However, solving these equations is notoriously difficult and the state-of-the-art methods are grid-based, limiting their application in higher dimensional problems~\cite{bansal2017hamilton, nagami2021hjb}.
Figure~\ref{fig:teaser} illustrates a high-level comparison of these approaches.
Numerical methods that use a grid to discretize the state space are limited to problems in lower dimensions. On the other hand, local trajectory optimization methods compute control laws from a single initial state to reach the target.
Our method provides a compromise between these two approaches. Similar to local methods, it is grid-free and can be used in higher dimensional problems. Different from them, our method can generate control laws starting from initial states sampled from a large portion of the state space.

To this end, our contributions are two-fold: 
1) We propose to approximate the state transitions using neural networks with sinusoidal activation functions and supervise them with numerically computed gradients of the system dynamics, 
2) We present a method that integrates the learned dynamics into the HJB equations which are used to train the networks representing the controller and the value function. In our experiments, we show that our method can be used in a variety of systems in a sample efficient fashion.

\section{Related Work}

In this section, we review related work on various approaches for solving nonlinear optimal control problems and their relationship with our proposed method.
Optimal control is concerned with finding a control law to drive a dynamical system to a goal state, such that a given cost function is minimized while satisfying the physical constraints of the system. 
Due to their broad applicability, there are many approaches for solving optimal control problems~\cite{rao2009survey}. 
While systems with linear first-order constraints have closed-form optimal controllers~\cite{bemporad2002explicit, basin2005closed}, their nonlinear counterparts often do not have analytic solutions~\cite{tang2019data}. Consequently, numerical methods have been commonly used to compute the controller for nonlinear systems~\cite{rao2009survey}.

Dynamic programming is a useful tool to efficiently compute optimal solutions in discrete time and space, using the Bellman equation.
The continuous time analog of the Bellman equation is the HJB equation, a partial differential equation (PDE) in $d$ state dimensions plus time. The solution of this PDE can be used to generate control inputs from arbitrary initial conditions from the state space. However, while offering global control laws, its computation can be costly when discretization of the state space is required to solve the PDE. This problem is commonly known as the curse of dimensionality. There are many studies for solving the HJB PDEs using numerical methods, such as level set methods~\cite{sethian1996fast} and the essentially nonoscillatory scheme~\cite{osher1991high, zhang2016eno}. Yet, even the state-of-the-art numerical methods are limited to state spaces of dimension 4 to 5, since they rely on grid-based discretization of the state space~\cite{bansal2017hamilton}. Numerical methods based on HJB PDEs were shown to control Dubins vehicles in cluttered environments~\cite{takei2010practical, parkinson2020hamilton} and in a differential game setting~\cite{mitchell2005time}.

Nonlinear Model Predictive Control (NMPC) is another strategy to solve nonlinear optimal control problems. NMPC predicts the behavior of a system over some time horizon, such that the control input to the system optimizes the given cost function~\cite{findeisen2002introduction}. This approach uses standard optimization techniques~\cite{diehl2009efficient} to solve the optimization problem starting from the initial state, and does not require space discretization. NMPCs have been successfully used for controlling nonlinear systems with complex dynamics, such as quadrotors~\cite{zanelli2018nonlinear, wang2021efficient} and fixed-wing aircrafts~\cite{kang2009linear, stastny2018nonlinear}. However, since the optimization is performed along trajectories starting from a single initial state, the controller needs to be recomputed for each new initial condition. Moreover, the original formulation can require an accurate model of the dynamics to effectively control the system~\cite{bemporad1999robust, lopez2019dynamic}.
An alternative to NMPC is continuous-time Model-based Reinforcement Learning (CT-MBRL), which enables learning policies to solve complex tasks through learning the system's model and the policy for the agent~\cite{doya2000reinforcement, yildiz2021continuous}.
Model-based reinforcement learning methods can improve the sample complexity compared to model-free approaches~\cite{moerland2020model, deisenroth2011pilco, gal2016improving, song2021pc, yildiz2021continuous}. 
Nevertheless, for simple control problems these model-based methods still require sizable interaction data during training to learn optimal policies compared to HJB-based neural controllers, as we show in our experiments (Section~\ref{sec:exp-learnedfk}).

Recently, the advances in using neural networks to solve problems involving differential equations~\cite{chen2018neural, raissi2019physics, sitzmann2020implicit}  led to an increasing interest in computing HJB-based controllers in a grid-free fashion, by learning the value function.
One way to learn the value function is through direct supervision on the cost-to-go values~\cite{huh2021learning}.
If such supervision is not available, the existing HJB-based neural optimal control methods require the actual dynamics equations to analytically compute the Hamiltonian that is used to learn the value function and generate the control signals~\cite{onken2021neural, bansal2021deepreach, lutter2020hjb}.
Our method builds on this line of work and uses a neural network to approximate the value function. However, different from these works, it uses learned system dynamics to integrate the outputs of the controller, and supervises the control law with the HJB PDEs computed using the learned dynamics.
This is in contrast to existing approaches that derive the optimal controller analytically from the Hamiltonian based on known dynamics, whose solution requires special care when the control input is multi-dimensional and has saddle points.
Additionally, our learned controller generates entire trajectories in one go instead of finding the optimal control based on the Hamiltonian at each time step, which allows fast training time performance.

\approaches

Table~\ref{tab:approaches} summarizes the comparison of different approaches to solve nonlinear optimal control problems in terms of whether the approach \emph{i)} provides a global solution, \emph{ii)} requires space discretization, \emph{iii)} requires the true dynamics of the system, and \emph{iv)} has good sample efficiency.
Our method can generate control laws from multiple initial states that lie in a large portion of the state space, while not requiring space discretization and the true state transitions. In addition, it has better efficiency compared to recent MBRL methods. 

\section{Problem Statement}

Our problem statement has a similar form to the classical optimal control formulations.
For a fixed and finite time-horizon $[t_0, t_f]$, the state evolution is given by the following ordinary differential equation
\begin{equation}
\label{eq:state_evo}
    \dot{\mymat{x}}(t) = f(\mymat{x}(t), \mymat{u}(t))
\end{equation}

where $\mymat{x}(t) \in \mathcal{X} \subseteq \mathbb{R}^d$ and $\mymat{u}(t) \in \mathcal{U} \subseteq \mathbb{R}^m$ denote the state and action at time $t \in [t_0, t_f]$, respectively.
The state space of the system is $\mathcal{X}$, and the constrained action space is denoted by $\mathcal{U}$.
Given an initial state $\mymat{x}_0$ at time $t_0$, we have a continuous-time cost functional
\begin{equation}
    J(\mymat{x}(t_0), \mymat{u}(\cdot), t_0) = \int_{t_0}^{t_f} L(\mymat{x}(t), \mymat{u}(t), t) dt + G(\mymat{x}(t_f))
\end{equation}

subject to the first-order constraints given in Eq.~\ref{eq:state_evo}, action constraints $\mymat{u}(t) \in \mathcal{U}$, and $\mymat{x}(t_0) = \mymat{x}_0$. 
Here, $L: \mathbb{R}^d \times \mathbb{R}^m \times [t_0, t_f] \rightarrow \mathbb{R}$ is the running cost, whereas $G: \mathbb{R}^d  \rightarrow \mathbb{R}$ is the terminal cost. The cost functional $J$ and the dynamics $f$ are assumed to be continuously differentiable in $\mymat{x}$ and $\mymat{u}$.
Assuming its existence, our goal is to find the optimal controller $\mymat{u}^*(t), \forall t \in [t_0, t_f]$, such that the cost functional is minimized:
\begin{equation}
    \mymat{u}^*(\cdot) = \arg\min_{\mymat{u}(\cdot) \in \mathcal{U}}{J(\mymat{x}(t), \mymat{u}(\cdot), t)}
\end{equation}
which generates the optimal state trajectory.
The optimal controller also yields the value function $V: \mathbb{R}^d \times [t_0, t_f] \to \mathbb{R}$, the optimal cost starting from $(\mymat{x}(t), t)$, 
\begin{equation}
    V(\mymat{x}_t, t) = \min_{\mymat{u}_{t:}}{J(\mymat{x}_t, \mymat{u}_t, t)}
\end{equation}
Here and in the remainder of the paper, we use the $\mymat{x}_t$ and $\mymat{u}_t$ notation for the state and control signal at time $t$. We additionally use the notation $\mymat{u}_{t:}$ to indicate the control in the time interval $[t, t_f]$, and $\mymat{u}_{:}$ for the entire interval $[t_0, t_f]$. Therefore, $\mymat{x}^*_:$ denotes the optimal state trajectory and $\mymat{u}^*_:$ is the optimal control.


\section{Background}

In this section, we review the Pontryagin's Maximum Principle (PMP) and the Hamilton-Jacobi-Bellman (HJB) equations that our method builds on.
The PMP provides the necessary first-order conditions for the optimality of a controller.
First, we need the control Hamiltonian which is given by
\begin{equation}
\label{eq:hamiltonian}
    H(\mathbf{x}_t, \lambda_t, t) = \min_{\mathbf{u}_t} \big\{L(\mathbf{x}_t, \mathbf{u}_t, t) + \lambda_t^\top \cdot f(\mathbf{x}_t, \mathbf{u}_t, t)\big\}
\end{equation}

where $\lambda_t$ is a vector of costate variables defined by the gradient of the value function with respect to the states, $\lambda_t = \nabla_{\mymat{x}} V(\mymat{x}_t, t)$\footnote{We use the notation $\nabla_\mathbf{x}$ for gradients w.r.t. a vector $\mathbf{x}$, and $\timederiv \mathbf{x} / \partial t$ or $\dot{\mathbf{x}}$ for time derivatives of $\mathbf{x}$.}. 
Then, according to the PMP, the following conditions for optimality are necessary at all times $t$:
\begin{align}
    \timederiv \mathbf{x}^*_{t} /\partial t &= \nabla_\lambda H(\mathbf{x}^*_t, \lambda^*_t, t) \nonumber\\
    \timederiv \lambda^*_{t} /\partial t &= -\nabla_\mathbf{x} H(\mathbf{x}^*_t, \lambda^*_t, t) \\
    \mathbf{0} &= \nabla_\mathbf{u} H(\mathbf{x}^*_t, \lambda^*_t, t) \nonumber
\end{align}

This means that once the value function $V$ and its gradients $\nabla_{\mymat{x}} V$ are known, the optimal control signals can be computed from any point from the state space.
The value function satisfies the HJB equation, which involves the partial derivatives of $V$ with respect to $t$ and $\mymat{x}$ as follows, 
\begin{equation}
    \partial V(\mymat{x}_t, t) / \partial t + H(\mathbf{x}_t, \lambda_t, t) = 0
\end{equation}

with the boundary condition $V(\mymat{x}_f, t_f) = G(\mymat{x}_f)$. The HJB equation is derived using the principle of optimality.
We refer the readers to~\cite{kirk2004optimal, anderson2007optimal, liberzon2011calculus} for their derivations.

\section{Method}



In this section, we describe our method for learning the dynamics of the system from data, and using the learned system dynamics to train a controller function.

\subsection{Learning State Transitions from Data}
\label{sec:fklearning}

Our method takes a model-based approach for learning the controller that minimizes the cost functional.
To do so, in an offline process, we train a neural network $f_\theta$ that approximates the state transitions $\dot{\mymat{x}}_t = f(\mymat{x}_t, \mymat{u}_t)$.


We use a multi-layer perceptron (MLP) $f_\theta: \mathbb{R}^d \times \mathbb{R}^m \to \mathbb{R}^d$ with parameters $\theta$ and sine ($\sin$) activations to represent the state transition function, as inspired by~\cite{sitzmann2020implicit} who showed the success of periodic activation functions for fitting images and 3D shapes.
To train $f_\theta$, we first collect a dataset $\mathcal{D} = \{\mymat{x}_i, \mymat{u}_i\}_{i=1}^N$ of state and action pairs that are sampled uniformly from the state and action spaces.
We use the ground truth transitions $f$ to generate the desired outputs $\dot{\mymat{x}}$ and train the network in a supervised fashion.


In our method, we will use the learned transition function $f_\theta$ to compute the Hamiltonian with Eq.~\ref{eq:hamiltonian}, and optimize the controller with the Hamiltonian's gradients.
Therefore, in addition to the state derivatives $\dot{\mymat{x}}$, we use the gradients of the transition function with respect to $\mymat{x}$ and $\mymat{u}$, $\nabla_\mymat{x} f$ and $\nabla_\mymat{u} f$, as supervision.
Since the nonlinearities of $f_\theta$ are sinusoidal functions the gradients of the network are well-defined, and the network can approximate the transitions effectively.
We use gradients of the state transitions computed using automatic differentiation to supervise the derivatives of $f_\theta$ with respect to $\mymat{x}$ and $\mymat{u}$.

The loss function we use for system identification is 
\begin{align*} 
\mathcal{L}_\text{sys-id} &= \textstyle\sum_i \Vert f_\theta(\mymat{x}_i, \mymat{u}_i) - f(\mymat{x}_i, \mymat{u}_i) \Vert \\ 
    &+ \Vert \nabla f_\theta(\mymat{x}_i, \mymat{u}_i) - \nabla f(\mymat{x}_i, \mymat{u}_i) \Vert
\end{align*}
where $\nabla f$ includes derivatives w.r.t. both $\mymat{x}$ and $\mymat{u}$.

Concurrent to our work,~\cite{kim2022learning} recently showed the advantage of supervision with function gradients to learn the state transitions of a manipulator with 6 degrees of freedom (DoF). Unlike this study, our approach uses networks with sinusoidal activations which leads to better identification performance for the systems we study, as shown in our experiments.

\begin{figure}
    \centering
    \includegraphics[width=\columnwidth]{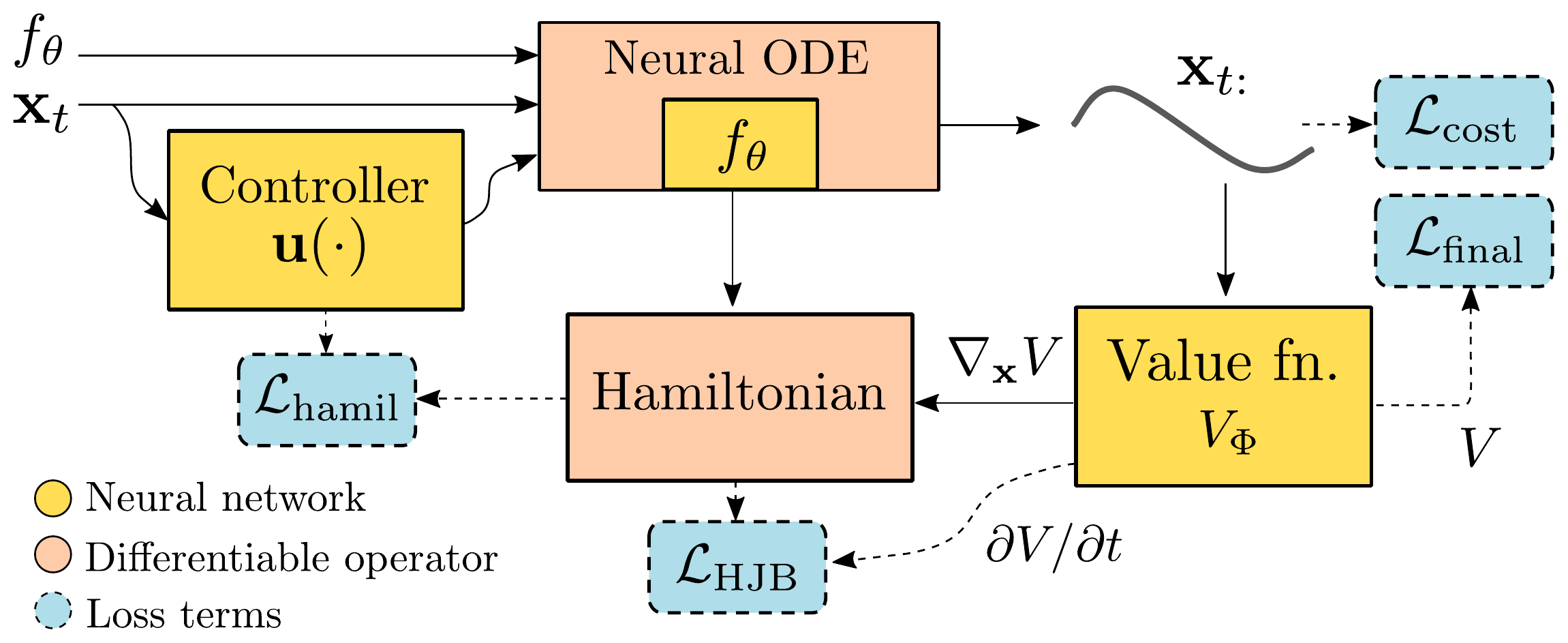}
    \caption{\textbf{Method overview.} Our method takes as input a set of initial states $\mymat{x}_t$, and a learned state transition function $f_\theta$. The outputs of the controller $\mymat{u}(\cdot)$ are integrated by a Neural ODE using $f_\theta$ to generate the state trajectories $\mymat{x}_{t:}$ in a feedback loop. We use loss functions adapted by the HJB equation to train the controller and the value function $V_\Phi$ jointly. At test time, to evaluate the learned controller we use the actual state evolution function $f$.}
    \label{fig:method}  
\end{figure}

\subsection{The Value Function and Controller Design}

We propose a learning-based strategy to generate controllers for solving control problems. Our method, which we call Control with Neural Dynamics (or \optctrl{}), uses neural networks for representing both the value function and the controller. An overview of our method is shown in Figure~\ref{fig:method}. 

Starting from a set of initial states sampled from a distribution, \optctrl{} computes state trajectories using the learned state transitions $f_\theta$ (Section~\ref{sec:fklearning}) along with the control signals generated by the controller. 
The parameters of the controller $\mathbf{u}(\cdot)$ are initialized uniformly at random, leading to state trajectories initially with high cost.
We train the controller and the value function jointly by minimizing the cost functional and violations of the HJB equations.

The controller $\mathbf{u}(\cdot)$ takes the state at the current time $\mymat{x}_t$ as input, and outputs the control signal $\mymat{u}_t$. The $\mathbf{u}(\cdot)$ function is designed to be an MLP whose last layer is followed by a hyperbolic tangent $(\tanh)$ and appropriate rescaling depending on the system, to constrain the outputs be within the action space $\mathcal{U}$. 
We assume the stability of the system as long as $\mymat{u}_t \in \mathcal{U}, \forall t \in [t_0, t_f]$, starting from a viable state $\mymat{x}_0$. 
The controller outputs $\mymat{u}_t$ are integrated in conjunction with $f_\theta$ by a neural ordinary differential equation (ODE) solver~\cite{chen2018neural} to generate the state trajectory $\mymat{x}_{t:}$ in a feedback loop.

\optctrl{} represents the value function $V_\Phi: \mathbb{R}^d \times [t_0, t_f] \to \mathbb{R}$ also by an MLP with $\tanh$ activations and skip connections, in a similar vein with~\cite{onken2021neural}. 
Approximating the value function using neural networks has the advantage of being grid-free, which means that the space complexity does not grow exponentially with the state dimension. Consequently, we are able to use our method for higher dimensional systems which may not be possible with existing numerical solvers.

The outputs of the value function $V_\Phi(\mymat{x}_t, t)$, as well as its derivatives w.r.t. $\mymat{x}$ and $t$ are used to compute the Hamiltonian and for supervising the controller outputs using constraints from the HJB equations.
The Hamiltonian is computed as
\begin{equation}
    H = L(\mathbf{x}_t, \mathbf{u}_t, t) + \nabla_\mymat{x} V_\Phi(\mymat{x}_t, t) ^\top \cdot f_\theta(\mathbf{x}_t, \mathbf{u}_t, t)
\end{equation}
where $L$ is the known running cost function, and $\mathbf{x}_t$ are the states dictated by the controller outputs $\mathbf{u}_t$.

In contrast to recent neural optimal control studies, such as~\cite{onken2021neural, bansal2021deepreach}, our method cannot analytically derive the Hamiltonian and the optimal controller for each system, since the state transitions are assumed to be unknown. Instead, we use the loss functions described in the next section to optimize the controller and the value function.

\subsection{Loss Functions and Training Details}

Our goal is to train the value function and the controller for learning to generate optimal control signals.
In our setting, there is no expert strategy or labeled controller data. Therefore, we opt for using loss functions adapted from the HJB PDEs. 
The first loss function we use optimizes the controller outputs by penalizing the running and terminal costs of the entire state trajectory,
\begin{equation}
    \mathcal{L}_\text{cost} = \textstyle \mathbb{E}_{\mymat{x}_0 \sim \rho} \int_{t_0}^{t_f} L(\mathbf{x}_t, \mathbf{u}_t, t) dt + G(\mathbf{x}_f)
\end{equation}
where $\rho$ is the distribution the initial states are sampled from.
Additionally, we optimize the value function for each state along the trajectory by
\begin{equation}
\mathcal{L}_\text{HJB} = \Vert \partial V_\Phi(\mymat{x}_t, t) / \partial t + H \Vert_1
\end{equation}
and the boundary condition at the final time
\begin{equation}
\mathcal{L}_\text{final} = \Vert V_\Phi(\mathbf{x}_f, t_f) - G(\mathbf{x}_f) \Vert_1
\end{equation}

Finally, we use the gradients of the Hamiltonian w.r.t. $\mymat{u}$ to regularize the controller
\begin{equation}
\mathcal{L}_\text{hamil} = \Vert \nabla_\mymat{u} H \Vert
\end{equation}

The total loss is then a weighted sum of these loss terms:
\begin{equation}
\mathcal{L} = \alpha_\text{cost} \mathcal{L}_\text{cost} + \alpha_\text{HJB} \mathcal{L}_\text{HJB} + \alpha_\text{final} \mathcal{L}_\text{final} + \alpha_\text{hamil} \mathcal{L}_\text{hamil} 
\end{equation}
where the weights are set to $\alpha_\text{cost} = 1, \alpha_\text{HJB} = 1, \alpha_\text{final} = 0.01, \alpha_\text{hamil} = 0.01$ in our experiments.
We use the Adam optimizer~\cite{kingma2014adam} with exponentially decaying learning rate starting from 0.01 to train the networks.

\section{Experiments}
In this section, we present experimental results comparing our method against the state-of-the-art.
We design our experiments for investigating the following. 

\begin{enumerate}
\item Design choices for the network and its supervision for learning the state transitions of a system
\item Optimal control performance for systems whose state transitions are not known beforehand
\item Optimal control performance for systems with known state transitions
\end{enumerate}

In our experiments, we use recent MBRL algorithms, a neural optimal controller and a sampling-based motion planning algorithm to compare against our method.

    

\subsection{Learning state transitions from data}

\fklearning

In our first experiment, we analyze the performance of different strategies to approximate the state transitions $\dot{\mymat{x}}_t = f(\mymat{x}_t, \mymat{u}_t)$ with a neural network $f_\theta$, which is used as part of our method to learn the controller.

We compare strategies using different activation and loss functions to learn the state transitions of four systems: Dubins car, Acrobot, Cartpole and Quadrotor.
Two of these, Acrobot and Cartpole, are well-studied systems that appear in benchmarks evaluating control and reinforcement learning algorithms~\cite{brockman2016openai, tassa2018deepmind}.

The other two systems, the Dubins car and Quadrotor, are commonly used motion models for robotics problems.
The differential equation modeling the state evolution of a Dubins car with varying speed is
\begin{equation}
\label{eq:dubins}
\dot{\mymat{x}}_t = [\dot{x}_t, \; \dot{y}_t, \; \dot{\psi}_t]^\top = [v_t \cos(\psi_t), \; v_t \sin(\psi_t), \; \alpha_t \cdot v_t / r]^\top
\end{equation}

where $(x_t, y_t)$ denotes the position of the robot and $\psi_t$ is the heading angle.
The control signal includes the linear velocity $v_t \in [0, v_{max}]$ and the steering $\alpha_t \in [-1, 1]$. 

The quadrotor system is controlled by its four motors. The dimension of this system's state space is 12, including the 3D positions, orientations in Euler angles and their time derivatives.
We use the dynamics model given in~\cite{garcia2013modeling}.

Table~\ref{table:fk_learning} presents results comparing the design choices for the network architecture and its supervision.
We compare networks with the Rectified Linear Unit (ReLU), $\tanh$ and $\sin$ activations.
For each of these strategies, we use a three-layer MLP whose hidden units are followed by the corresponding activation functions.
We train each of these networks for 50,000 epochs using $(\mymat{x}_t, \mymat{u}_t)$  data sampled from the state and action spaces.
The loss function used to train these networks are either with or without the $\nabla f = [\nabla_\mymat{x} f, \nabla_\mymat{u} f]$ supervision which is computed using PyTorch’s auto-differentiation engine.

We report the average and median errors of the predicted state transitions $f_\theta(\mymat{x}_t, \mymat{u}_t)$.
We find that neural networks with sinusoidal activation functions outperform the other tested activation functions for identifying the state transitions in all four tasks.
This can be partly due to the trigonometric functions involved in all of these systems, and the angular shift-invariance that can be captured better with periodic activations.
Additionally, we observe the largest gains over the other nonlinearities in the systems of Acrobot and Quadrotor more so, which is the system with the highest state space dimension among them.
Using the $\nabla f$ supervision, we see further performance improvement in mean error for predicting the state transitions of the Dubins car, Acrobot and Cartpole systems. On the Quadrotor, however, it degrades the performance for all the activation functions. This may be attributed to the scale difference between the entries of the Jacobian, causing instabilities during training.
Nevertheless, the performance of the sinusoidal activations for learning the state transitions is effective, and we use the learned function $f_\theta$ for generating the controllers in the following section.

\subsection{Optimal control using learned dynamics}
\label{sec:exp-learnedfk}

Next, we evaluate our method for generating a controller using learned dynamics.
We compare our method against approaches from two lines of works: model-based reinforcement learning (MBRL) algorithms that learn the transitions model while optimizing the policy, and a recent neural optimal control method that derives the control law assuming known dynamics.

In our experiments, the running cost of the cost functional for our method is set to $(\mymat{u}_t - \mymat{u}^*)^\top R (\mymat{u}_t - \mymat{u}^*)$, and the terminal cost is $(\mymat{x}_t - \mymat{x}^*)^\top P (\mymat{x}_t - \mymat{x}^*)$.
Here the matrices $R$ and $P$ are used to weight the relative importance of the cost terms. Unless noted otherwise, we use $P = I_{d}$ and $R = 0.01 I_{m}$, where $I$ is the identity matrix.
The desired state and action, $\mymat{x}^*$ and $\mymat{u}^*$, are given in all the methods.




The first experiment compares our method against model-based reinforcement learning methods Deep PILCO~\cite{gal2016improving}, PETS~\cite{chua2018deep} and ENODE~\cite{yildiz2021continuous}.
These MBRL methods are designed to model the transition probabilities (or the state transitions) using different approaches. Deep PILCO uses Bayesian neural networks to approximate the transitions, whereas PETS uses an ensemble of probabilistic neural networks. The most recent of them, ENODE, is designed to solve continuous time control problems by approximating the system dynamics with an ensemble of neural ODEs.

The results comparing our method against the MBRL baselines are presented in Table~\ref{table:optctrl_gym}, where we report the terminal error, the control inputs applied over the course of the trajectory and the number of evaluations of the learned dynamics $f_\theta$ for forward simulation during training.
We found that Deep PILCO outperforms the other MBRL methods in both the Acrobot and Cartpole tasks. Our method's performance, on the other hand, is comparable to the baselines in the Acrobot environment and is better in the Cartpole task.
Notably, our method requires significantly fewer number of evaluations of the learned transition function during training.
This leads to much faster training time performance- whereas training the baselines take more than an hour on a Tesla V100, our method takes about 15 minutes. 

\gymtasks

In the second experiment, we compare against NeuralOC~\cite{onken2021neural}, a HJB-based neural optimal controller that our method builds on. Similar to our method, NeuralOC trains the value function using penalty terms that enforce the HJB PDEs.
However, unlike our method it requires the true state transitions $f$, and uses $f$ to analytically derive the control law for a given system.

\begin{figure}[ht]
    \centering
    \includegraphics[width=\columnwidth]{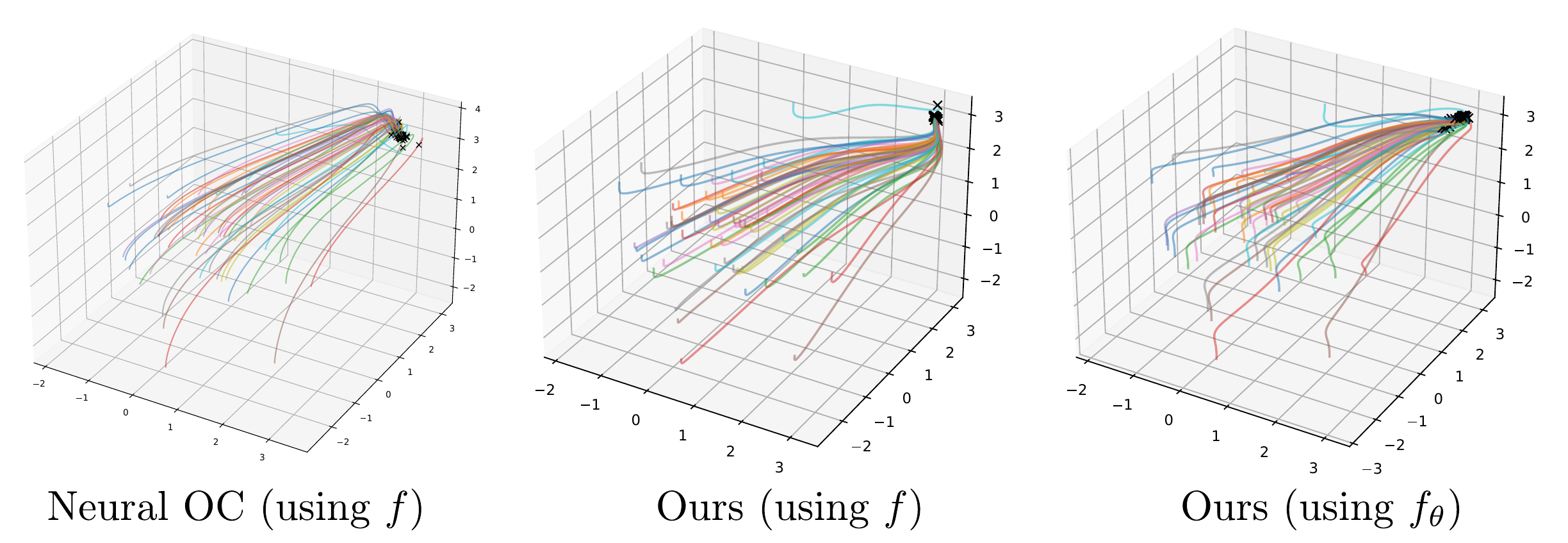}
    \caption{Trajectories generated by NeuralOC and our method for controlling a quadrotor to reach a desired state at $[3,3,3]$.}
    \label{fig:quadrotors} 
\end{figure}

We evaluate these methods on the task of controlling a quadrotor to reach a goal pose.
The goal pose is set to the position $[3,3,3]$ in an upright orientation.
The initial state positions are sampled from a normal distribution around the origin $\mathcal{N}([0,0,0], I)$, and the rest of the state variables are initialized as 0, corresponding to an upright orientation and no initial velocities. Table~\ref{table:quadrotor} presents quantitative results where we evaluate the methods starting from 1000 different initial states. Similar to the previous setting, we report the mean terminal state error and magnitude of control signals that generate the trajectories. To train NeuralOC, we use the hyperparameters given in~\cite{onken2021neural} for the quadcopter task as is, except for the distribution the initial states are sampled from, which we enlarged. 
In terms of the terminal state error, we find that our method significantly outperforms NeuralOC if the transition function $f$ is known. If the state transitions (S.T.) are unknown a priori, our method using the learned dynamics $f_\theta$ performs worse, but competitively. Remarkably, when we set the weight of the control input cost $R$ to zero, our method trained with imperfect dynamics performs better than the baseline.
Compared to all of our methods, we see that NeuralOC consumes the least energy on the control signals. Whereas our methods tend to maximize the control magnitude as long as they are within the action space.
We observe this phenomenon qualitatively as well, as shown in Figure~\ref{fig:quadrotors}.
The trajectories generated by NeuralOC tilt towards the goal position and start moving there, while our method trained with $f$ learns to first go to a location under the goal state position and move upwards for better terminal state error performance. 
The controls generated by our method trained with $f_\theta$, on the other hand, initially moves the quadrotor up and then move towards the goal for optimizing the terminal error at the expense of exerting larger control inputs.

\quadrotorctrl

\subsection{Optimal control using known dynamics}

\begin{figure}[h]
    \centering
    \includegraphics[width=.8\columnwidth]{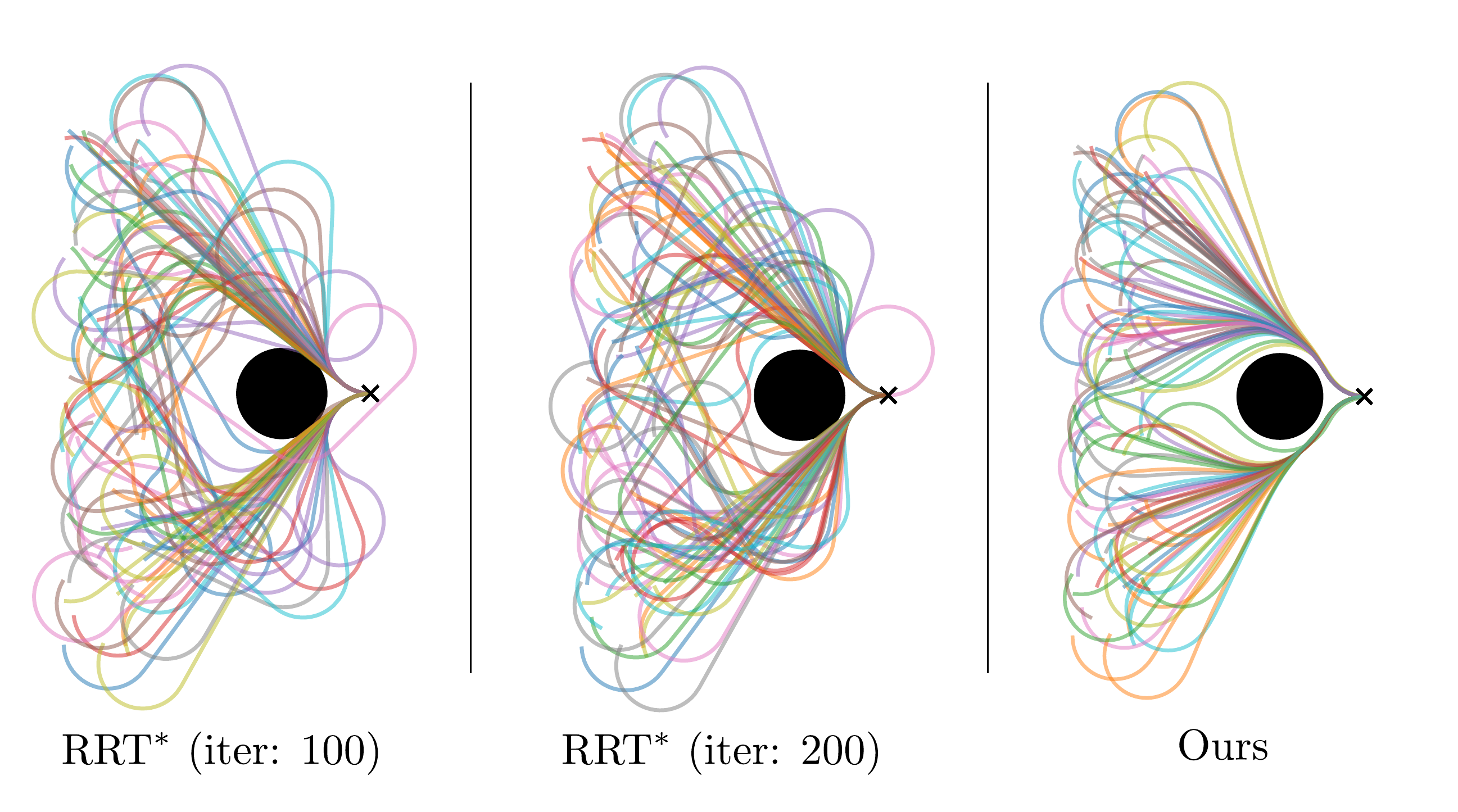}
    \caption{Trajectories generated by RRT$^*$ and our method for controlling a Dubins car to reach a desired state in the presence of a circular obstacle.}
    \label{fig:dubins} 
\end{figure}

In our final experiment, we evaluate the performance of our controller when it is trained with known state transitions. 
We use the Dubins car with equations of motion given by Eq.~\ref{eq:dubins}, in an environment containing a circular obstacle. 
We compare our method against RRT$^*$~\cite{karaman2011sampling}, a variant of the rapidly-exploring random tree which is a sampling-based motion planning algorithm.
Since this baseline is not a controller, but a planner, the concepts of terminal state error and magnitude of applied control are not applicable. Therefore, we instead compare the path lengths returned by  RRT$^*$ and the length of the trajectories our method computes. 
The initial states in this experiment are sampled from a uniform distribution with lower bound $[-3.5, -3, -\pi]$ and upper bound $[-2.5, 3, \pi]$. The goal state is $[0, 0, 0]$ and the circular obstacle of radius $0.5$ is located at $(-1, 0)$.
We use the RRT$^*$ implementation from the PythonRobotics library~\cite{sakai2018python}.
The collision avoidance objective is incorporated into our controller's cost function as a penalty term, similar to~\cite{onken2021neural, lutter2020hjb}.
Table~\ref{table:optctrldubins} compares the performance of our method against RRT$^*$.
Since it is an iterative algorithm whose solutions improve over time, we report two different versions of RRT$^*$ with different thresholds on the maximum number of iterations, to show how the computation time is compromised to improve the output paths.
We find that our method generates trajectories with better quality in terms of the path length. Additionally, its computation time during evaluation is two orders of magnitude smaller and has better overall success rate.
Figure~\ref{fig:dubins} shows the qualitative performance of each method. We see that with RRT$^*$ there are unnecessary detours resulting in suboptimal output paths.

\dubinsctrl

\section{Conclusion}
In this paper, we presented a method to generate control signals to drive a system to its desired state without a known dynamics model. 
Our method learns the state transitions to forward simulate the trajectories by fitting a neural network. Then, it uses the learned transitions along with loss functions that enforce the HJB PDEs to supervise the controller.
We evaluated our method on a number of systems to demonstrate its global nature and sample efficiency.
Our method assumes that the state transitions can be learned in one shot, as opposed to MBRL methods that optimize the model and the policy repeatedly until convergence. While our approach is sufficient for the control problems studied in this paper, for more complex tasks such as navigation in uneven terrains~\cite{wei2021predicting}, dexterous manipulation~\cite{nagabandi2020deep} and legged locomotion~\cite{sun2021online}, continual optimization of the model may be necessary.
Additionally, our method learns the controller for a fixed goal state, and requires re-optimization for new goal states, which we plan to address in future work.

\vtxt{looks good!}


\bibliography{references}

\begin{thebibliography}{10}

\bibitem{johansen2011introduction}
Tor~A Johansen.
\newblock Introduction to nonlinear model predictive control and moving horizon
  estimation.
\newblock {\em Selected topics on constrained and nonlinear control}, 1:1--53,
  2011.

\bibitem{bansal2017hamilton}
Somil Bansal, Mo~Chen, Sylvia Herbert, and Claire~J Tomlin.
\newblock Hamilton-jacobi reachability: A brief overview and recent advances.
\newblock In {\em 2017 IEEE 56th Annual Conference on Decision and Control
  (CDC)}, pages 2242--2253. IEEE, 2017.

\bibitem{nagami2021hjb}
Keiko Nagami and Mac Schwager.
\newblock Hjb-rl: Initializing reinforcement learning with optimal control
  policies applied to autonomous drone racing.
\newblock In {\em Robotics: Science and Systems}, 2021.

\bibitem{rao2009survey}
Anil~V Rao.
\newblock A survey of numerical methods for optimal control.
\newblock {\em Advances in the Astronautical Sciences}, 135(1):497--528, 2009.

\bibitem{bemporad2002explicit}
Alberto Bemporad, Manfred Morari, Vivek Dua, and Efstratios~N Pistikopoulos.
\newblock The explicit linear quadratic regulator for constrained systems.
\newblock {\em Automatica}, 38(1):3--20, 2002.

\bibitem{basin2005closed}
Michael Basin and Jesus Rodriguez-Gonzalez.
\newblock A closed-form optimal control for linear systems with equal state and
  input delays.
\newblock {\em Automatica}, 41(5):915--920, 2005.

\bibitem{tang2019data}
Gao Tang and Kris Hauser.
\newblock A data-driven indirect method for nonlinear optimal control.
\newblock {\em Astrodynamics}, 3(4):345--359, 2019.

\bibitem{sethian1996fast}
James~A Sethian.
\newblock A fast marching level set method for monotonically advancing fronts.
\newblock {\em Proceedings of the National Academy of Sciences},
  93(4):1591--1595, 1996.

\bibitem{osher1991high}
Stanley Osher and Chi-Wang Shu.
\newblock High-order essentially nonoscillatory schemes for hamilton--jacobi
  equations.
\newblock {\em SIAM Journal on numerical analysis}, 28(4):907--922, 1991.

\bibitem{zhang2016eno}
Y-T Zhang and C-W Shu.
\newblock Eno and weno schemes.
\newblock In {\em Handbook of Numerical Analysis}, volume~17, pages 103--122.
  Elsevier, 2016.

\bibitem{takei2010practical}
Ryo Takei, Richard Tsai, Haochong Shen, and Yanina Landa.
\newblock A practical path-planning algorithm for a simple car: a
  hamilton-jacobi approach.
\newblock In {\em Proceedings of the 2010 American Control Conference}, pages
  6175--6180. IEEE, 2010.

\bibitem{parkinson2020hamilton}
Christian Parkinson, Andrea~L Bertozzi, and Stanley~J Osher.
\newblock A hamilton-jacobi formulation for time-optimal paths of rectangular
  nonholonomic vehicles.
\newblock In {\em IEEE Conference on Decision and Control (CDC)}, pages
  4073--4078. IEEE, 2020.

\bibitem{mitchell2005time}
Ian~M Mitchell, Alexandre~M Bayen, and Claire~J Tomlin.
\newblock A time-dependent hamilton-jacobi formulation of reachable sets for
  continuous dynamic games.
\newblock {\em IEEE Transactions on automatic control}, 50(7):947--957, 2005.

\bibitem{findeisen2002introduction}
Rolf Findeisen and Frank Allg{\"o}wer.
\newblock An introduction to nonlinear model predictive control.
\newblock In {\em 21st Benelux meeting on systems and control}, volume~11,
  pages 119--141. Citeseer, 2002.

\bibitem{diehl2009efficient}
Moritz Diehl, Hans~Joachim Ferreau, and Niels Haverbeke.
\newblock Efficient numerical methods for nonlinear mpc and moving horizon
  estimation.
\newblock In {\em Nonlinear model predictive control}, pages 391--417.
  Springer, 2009.

\bibitem{zanelli2018nonlinear}
Andrea Zanelli, Greg Horn, Gianluca Frison, and Moritz Diehl.
\newblock Nonlinear model predictive control of a human-sized quadrotor.
\newblock In {\em 2018 European Control Conference (ECC)}, pages 1542--1547.
  IEEE, 2018.

\bibitem{wang2021efficient}
Dong Wang, Quan Pan, Yang Shi, Jinwen Hu, et~al.
\newblock Efficient nonlinear model predictive control for quadrotor trajectory
  tracking: Algorithms and experiment.
\newblock {\em IEEE Transactions on Cybernetics}, 51(10):5057--5068, 2021.

\bibitem{kang2009linear}
Yeonsik Kang and J~Karl Hedrick.
\newblock Linear tracking for a fixed-wing uav using nonlinear model predictive
  control.
\newblock {\em IEEE Transactions on Control Systems Technology},
  17(5):1202--1210, 2009.

\bibitem{stastny2018nonlinear}
Thomas Stastny and Roland Siegwart.
\newblock Nonlinear model predictive guidance for fixed-wing uavs using
  identified control augmented dynamics.
\newblock In {\em International Conference on Unmanned Aircraft Systems
  (ICUAS)}, pages 432--442. IEEE, 2018.

\bibitem{bemporad1999robust}
Alberto Bemporad and Manfred Morari.
\newblock Robust model predictive control: A survey.
\newblock In {\em Robustness in identification and control}, pages 207--226.
  Springer, 1999.

\bibitem{lopez2019dynamic}
Brett~T Lopez, Jean-Jacques~E Slotine, and Jonathan~P How.
\newblock Dynamic tube mpc for nonlinear systems.
\newblock In {\em 2019 American Control Conference (ACC)}, pages 1655--1662.
  IEEE, 2019.

\bibitem{doya2000reinforcement}
Kenji Doya.
\newblock Reinforcement learning in continuous time and space.
\newblock {\em Neural computation}, 12(1):219--245, 2000.

\bibitem{yildiz2021continuous}
Cagatay Yildiz, Markus Heinonen, and Harri L{\"a}hdesm{\"a}ki.
\newblock Continuous-time model-based reinforcement learning.
\newblock In {\em International Conference on Machine Learning}, pages
  12009--12018. PMLR, 2021.

\bibitem{moerland2020model}
Thomas~M Moerland, Joost Broekens, and Catholijn~M Jonker.
\newblock Model-based reinforcement learning: A survey.
\newblock {\em arXiv preprint arXiv:2006.16712}, 2020.

\bibitem{deisenroth2011pilco}
Marc Deisenroth and Carl~E Rasmussen.
\newblock Pilco: A model-based and data-efficient approach to policy search.
\newblock In {\em International Conference on Machine Learning}, pages
  465--472. Citeseer, 2011.

\bibitem{gal2016improving}
Yarin Gal, Rowan McAllister, and Carl~Edward Rasmussen.
\newblock Improving pilco with bayesian neural network dynamics models.
\newblock In {\em Data-Efficient Machine Learning workshop, ICML}, volume~4,
  page~25, 2016.

\bibitem{song2021pc}
Yuda Song and Wen Sun.
\newblock Pc-mlp: Model-based reinforcement learning with policy cover guided
  exploration.
\newblock In {\em International Conference on Machine Learning}, pages
  9801--9811. PMLR, 2021.

\bibitem{chen2018neural}
Ricky~TQ Chen, Yulia Rubanova, Jesse Bettencourt, and David~K Duvenaud.
\newblock Neural ordinary differential equations.
\newblock {\em Advances in Neural Information Processing Systems}, 31, 2018.

\bibitem{raissi2019physics}
Maziar Raissi, Paris Perdikaris, and George~E Karniadakis.
\newblock Physics-informed neural networks: A deep learning framework for
  solving forward and inverse problems involving nonlinear partial differential
  equations.
\newblock {\em Journal of Computational physics}, 378:686--707, 2019.

\bibitem{sitzmann2020implicit}
Vincent Sitzmann, Julien Martel, Alexander Bergman, David Lindell, and Gordon
  Wetzstein.
\newblock Implicit neural representations with periodic activation functions.
\newblock {\em Advances in Neural Information Processing Systems},
  33:7462--7473, 2020.

\bibitem{huh2021learning}
Jinwook Huh, Daniel~D Lee, and Volkan Isler.
\newblock Learning continuous cost-to-go functions for non-holonomic systems.
\newblock In {\em IEEE/RSJ International Conference on Intelligent Robots and
  Systems (IROS)}, pages 5772--5779. IEEE, 2021.

\bibitem{onken2021neural}
Derek Onken, Levon Nurbekyan, Xingjian Li, Samy~Wu Fung, Stanley Osher, and
  Lars Ruthotto.
\newblock A neural network approach for high-dimensional optimal control
  applied to multiagent path finding.
\newblock {\em IEEE Transactions on Control Systems Technology}, 2022.

\bibitem{bansal2021deepreach}
Somil Bansal and Claire~J Tomlin.
\newblock Deepreach: A deep learning approach to high-dimensional reachability.
\newblock In {\em IEEE International Conference on Robotics and Automation
  (ICRA)}, pages 1817--1824. IEEE, 2021.

\bibitem{lutter2020hjb}
Michael Lutter, Boris Belousov, Kim Listmann, Debora Clever, and Jan Peters.
\newblock Hjb optimal feedback control with deep differential value functions
  and action constraints.
\newblock In {\em Conference on Robot Learning}, pages 640--650. PMLR, 2020.

\bibitem{kirk2004optimal}
Donald~E Kirk.
\newblock {\em Optimal control theory: an introduction}.
\newblock Courier Corporation, 2004.

\bibitem{anderson2007optimal}
Brian~DO Anderson and John~B Moore.
\newblock {\em Optimal control: linear quadratic methods}.
\newblock Courier Corporation, 2007.

\bibitem{liberzon2011calculus}
Daniel Liberzon.
\newblock {\em Calculus of variations and optimal control theory: a concise
  introduction}.
\newblock Princeton university press, 2011.

\bibitem{kim2022learning}
Youngho Kim, Hoosang Lee, and Jeha Ryu.
\newblock Learning an accurate state transition dynamics model by fitting both
  a function and its derivative.
\newblock {\em IEEE Access}, 10:44248--44258, 2022.

\bibitem{kingma2014adam}
Diederik~P Kingma and Jimmy Ba.
\newblock Adam: A method for stochastic optimization.
\newblock {\em arXiv preprint arXiv:1412.6980}, 2014.

\bibitem{brockman2016openai}
Greg Brockman, Vicki Cheung, Ludwig Pettersson, Jonas Schneider, John Schulman,
  Jie Tang, and Wojciech Zaremba.
\newblock Openai gym.
\newblock {\em arXiv preprint arXiv:1606.01540}, 2016.

\bibitem{tassa2018deepmind}
Yuval Tassa, Yotam Doron, Alistair Muldal, Tom Erez, Yazhe Li, Diego de~Las
  Casas, David Budden, Abbas Abdolmaleki, Josh Merel, Andrew Lefrancq, et~al.
\newblock Deepmind control suite.
\newblock {\em arXiv preprint arXiv:1801.00690}, 2018.

\bibitem{garcia2013modeling}
Luis~Rodolfo Garc{\'\i}a~Carrillo, Alejandro~Enrique Dzul~L{\'o}pez, Rogelio
  Lozano, and Claude P{\'e}gard.
\newblock Modeling the quad-rotor mini-rotorcraft.
\newblock In {\em Quad Rotorcraft Control}, pages 23--34. Springer, 2013.

\bibitem{chua2018deep}
Kurtland Chua, Roberto Calandra, Rowan McAllister, and Sergey Levine.
\newblock Deep reinforcement learning in a handful of trials using
  probabilistic dynamics models.
\newblock {\em Advances in Neural Information Processing Systems}, 31, 2018.

\bibitem{karaman2011sampling}
Sertac Karaman and Emilio Frazzoli.
\newblock Sampling-based algorithms for optimal motion planning.
\newblock {\em The International Journal of Robotics Research}, 30(7):846--894,
  2011.

\bibitem{sakai2018python}
Atsushi Sakai, Daniel Ingram, Joseph Dinius, Karan Chawla, Antonin Raffin, and
  Alexis Paques.
\newblock Python{R}obotics: a python code collection of robotics algorithms.
\newblock 2018.

\bibitem{wei2021predicting}
Minghan Wei and Volkan Isler.
\newblock Predicting energy consumption of ground robots on uneven terrains.
\newblock {\em Robotics and Automation Letters}, 7(1):594--601, 2021.

\bibitem{nagabandi2020deep}
Anusha Nagabandi, Kurt Konolige, Sergey Levine, and Vikash Kumar.
\newblock Deep dynamics models for learning dexterous manipulation.
\newblock In {\em Conference on Robot Learning}, pages 1101--1112. PMLR, 2020.

\bibitem{sun2021online}
Yu~Sun, Wyatt~L Ubellacker, Wen-Loong Ma, Xiang Zhang, Changhao Wang, Noel~V
  Csomay-Shanklin, Masayoshi Tomizuka, Koushil Sreenath, and Aaron~D Ames.
\newblock Online learning of unknown dynamics for model-based controllers in
  legged locomotion.
\newblock {\em IEEE Robotics and Automation Letters}, 6(4):8442--8449, 2021.

\end{thebibliography}
\bibliographystyle{unsrt}




\end{document}